# Measuring Sustainability Intention of ESG Fund Disclosure using Few-Shot Learning

Mayank Singh*
Analytics & Research Divison
*Fidelity Investments*
Bengaluru, India
mayank.singh@fmr.com

Nazia Nafis*
Computer Science
Indian Institute of Information Technology
Lucknow, India
mcs21004@iiitl.ac.in

Abhijeet Kumar
Analytics & Research Divison
*Fidelity Investments*
Bengaluru, India
abhijeet.kumar@fmr.com

Mridul Mishra
Analytics & Research Divison
*Fidelity Investments*
Bengaluru, India
mridul.mishra@fmr.com

*Abstract*—Global sustainable fund universe encompasses open-end funds and exchange-traded funds (ETF) that, by prospectus or other regulatory filings, claim to focus on Environment, Social and Governance (ESG). Challengingly, the claims can only be confirmed by examining the textual disclosures to check if there is presence of intentionality and ESG focus on its investment strategy. Currently, there is no regulation to enforce sustainability in ESG products space. This paper proposes a unique method and system to classify and score the fund prospectuses in the sustainable universe regarding specificity and transparency of language. We aim to employ few-shot learners (parameter efficient finetuning of large language models) to identify specific, ambiguous, and generic sustainable investment-related language. Additionally, we construct a ratio metric to determine language score and rating to rank products and quantify sustainability claims for US sustainable universe. As a by-product, we publish manually annotated quality training dataset on Hugging Face (ESG-Prospectus-Clarity-Category under cc-by-nc-sa-4.0) of more than 1K ESG textual statements, obtained via a data extraction pipeline from summary prospectuses of funds and ETF. The performance of the few-shot finetuning approach is compared with zero-shot models e.g., Llama-13B, GPT 3.5 Turbo etc. We found that prompting large language models are not accurate for domain specific tasks due to misalignment issues. The few-shot finetuning techniques outperform zero-shot models by large margins of more than absolute ~30% in precision, recall and F1 metrics on completely unseen ESG languages (test set). Overall, the paper attempts to establish a systematic and scalable approach to measure and rate sustainability intention quantitatively for sustainable funds using texts in prospectus. Regulatory bodies, investors, and advisors may utilize the findings of this research to reduce cognitive load in investigating or screening of ESG funds which accurately reflects the ESG intention.

*Keywords—Sustainable products, Few-shot Learning, Finetuning language models, ESG Language Analysis, Greenwashing.*

## I. INTRODUCTION

In recent years, Sustainable fund assets has witnessed substantial rise globally. According to Morningstar [1], global sustainable fund assets stood at USD 2.74 trillion from 7K funds at the end of March 2023. The current sustainable assets in U.S constitutes ~11% of global market from 638 funds [1]. Sustainable products are investments that are included or excluded based on their evaluation of environmental, social and governance practices.

Fund managers explain the process of incorporating ESG factors in investment strategy section of fund's prospectus. These claims are the only publicly available disclosures for investors to understand if it is impact fund, focused funds, or integration funds [2]. The lack of regulations in ESG products has led to greenwashing in funds, investor confusions, non-compliance to standards, rating manipulation etc. Recently, U.S. Security and Exchange Commission (SEC) charges BNY Mellon Investment advisor and Godman Sachs Asset Management for misstatements and failing to follow policies [3][4]. Australian SEC outline an article on "How to avoid greenwashing when offering or promoting sustainability-related products" in information sheet 271 [5]. This paper attempts to highlight several issues listed in the article for example vague terminology requiring further clarifying disclosures, inadequate explanations, failure to disclose influence over the benchmark index etc. Table 1. shows few examples from prospectus of such language.

There are numerous approaches for analyzing prospectus text from frequency-based statistics to classification-based methods [6] [7]. The major challenge with count-based text measures is twofold. Firstly, defining keywords exhaustively is subjective and challenging. Secondly, its inability to recognize concealed intention of the language. There are several questions posed from the investment strategy which needs to be acknowledged. Has the fund manager mentioned ESG screening process with specific standards. Do they want to reveal thresholds for exclusions? Are they utilizing proprietary ratings or 3rd party independent vendor? There could be several such intentions which would be difficult for rule-based methods to apprehend. Furthermore, Classification approaches suffers in label scarce settings due to the issue of collecting labelled training data.

Recent advancement in large language models has enabled researchers to categorize texts with zero shot pretrained models [8][9] or prompt-based instruction tuned models [10][11][12] directly which requires no training data. We observe issues while experimenting with both models. Majorly, zero shot classifiers experiences performance issues due to absence of specific domain knowledge and prompt-based models (open-source) is limited by alignment problem, hallucinations, robustness, varying outcome of prompts etc. Hence, this paper employs few-shot finetuning of language models for classification task [13][14] using small training dataset annotated manually. It shows promising results qualitatively by learning many ways of expressing transparency indicators as wells as vagueness.

The paper describes overall system (two classification models) and methods (scoring and ranking) to measure and rank sustainable funds universe on bounded metric. The research paper will make the following contributions to growing body of literature in sustainable investing:
- We demonstrate novel application of few-shot finetuned language models in training language classifiers to investigate ESG compliance in disclosures.
- We construct a ratio metric for calculating language score and hence provide language rating to each fund.
- We publish the manually annotated dataset (utilized for finetuning) publicly [15].

## II. RELATED WORK

There is considerable research on the topic on greenwashing for funds, but the language analysis of fund's disclosures has been less explored in literature. Min Yi Li et al. has attempted to identify greenwashing using by classifying prospectus into good or bad classes using keyword frequency and DistilBert classification approach [6]. The study was limited to 94 funds, and it concluded that keywords frequency was not strong indicator and DistilBert model could not converge. Angie et al. studied discrepancy between text and fundamental ESG measures of funds by comparing text scores measured by ESG related intensity, positioning, readability, tonality, and uniqueness to holding based ESG ratings. They concluded that fund flows respond strongly to text based ESG measures [7]. In a different work, Liying showed that ESG risk disclosure (in fund prospectuses) can be used for risk management purposes to mitigate the adverse effects of high ESG risk exposure [16]. The author developed ESG vs non-ESG classifiers for ESG sentences from investment strategy and Risk sections separately. Cara Beth in her work advocates for mandating a standard ESG disclosure framework to limit inconsistency and reducing the likelihood of potential greenwashing [17]. There have been several articles published on the issue of greenwashing but very few machine learning assisted research to address the issue [18][19][20][21].

Recently, with advent of large language models there had been ground breaking advancements in prompt based zero-shot and few shot models [10][11][12]. Moreover, with the recent few shot finetuning techniques like parameter efficient finetuning (PEFT) and sentence transformer finetuning (SetFit) have proven research to perform at par with full finetuning models with few sampled training data [13][14][22]. Rajdeep Sarkar et al. applied few shot approaches to legal text classification. The authors concluded that performance is much better than zero shot and requires much less manual annotation to develop the system when compared to fully supervised models [23]. The outline of rest of the paper is as follows. In Section 3, we describe proposed methodology including few-shot models and scoring construction. In Section 4, we penned data creation process. Section 5. shows experimental details and result analysis.

## III. PROPOSED METHODOLOGY

To measure sustainability intention and focus of the fund, we begin by grounding the process describing flow diagram. Further, the few shot finetuning models are discussed. Finally, we construct the scoring and rating method for final ranking of funds. Additionally, the overall system may assist ESG professionals to quickly analyse the language of interest reducing their cognitive load.

### A. ML System Design

We formulated a pipeline which reads the ESG prospectus data, follows a streamlined methods and models, and rates each sustainable product.

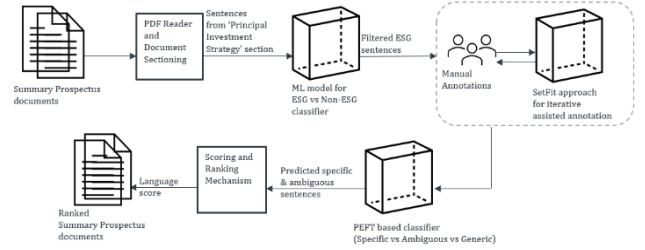

Figure 1: Process Design: Flow of data from methods & models.

Figure 1. depicts 5-steps process flow of proposed system with various modules. Initially, the disclosures are ingested into the system with the help of a PDF Reader and use of rule-based document sectioning techniques to automatically detect and extract the 'Principal Investment Strategy' sections (Step 1). Next, individual sentences from the 'Principal Investment Strategy' are extracted and passed through a Logistic Regression model (LR) to classify ESG vs. non-ESG sentences (Step 2). This step is performed to analyse the fund language vis a vis its ESG investment strategies only. Through experiments, it was established LR model was most cost effective without compromising performance when compared to finetuned large language model. Hence, there was no value in deploying LLM based classifier for ESG vs non-ESG classifier (Section 5).

Step 3 describes a SetFit model assisted iterative annotation process which required human in the loop for labelling specific intentions which builds credibility of sustainable funds. The necessity of data centric approach to collect quality data appeared due to inability of zero-shot large language models to categorize intentions correctly (check Section 4. for details). In step 4, The complete annotated dataset is used with a Parameter Efficient Fine-tuning (PEFT) approach known as Prompt Tuning to train the following BERT-family models: bert-base, roberta-base, and deberta-base [24][25][26], for the task of text clarity classification (the three classes being: Specific ESG sentences, Ambiguous ESG sentences, and Generic ESG sentences). Finally, the classified ESG sentences from its respective source documents are passed to scoring method to calculate the ESG language score for each document (step 5). The scoring methodology is explained in detail in Section 3.3. Based on the obtained scores, system objectively rates and rank all documents that exist within corpus to benefit the end users.

### B. Few-Shot Fine-Tuned Language Models
- SetFit is fine-tuning of a pretrained sentence transformers on a small number of text pairs, in a contrastive Siamese manner and is known to achieve high accuracy with orders of magnitude less parameters than the complete fine-tuning method [13][22]. We deploy SetFit based model to assist data annotation process. During experiments, three sentence transformer models, namely miniLM-v2, paraphrase-mpnet-v2, and sentence-t5-base were trained. Apart from being faster at inference and training, SetFit was performant even with small base models and did not require heavy compute.

- PEFT approaches work by fine-tuning only a small fraction of all model parameters, and thereby are more cost efficient than complete fine-tuning approaches [14]. Prompt Tuning makes use of soft prompts to condition frozen language models to perform the downstream tasks. Soft prompts are learned through back-propagation and can be tuned to incorporate signals from labelled examples. We perform Prompt Tuning of bert-base, roberta-base and deberta-base [24][25][26], as our final set of Parameter Efficient Fine-tuning (PEFT) experiments.

*C. Scoring & Ranking Method*

Motivation behind the constructing a language metric was to measure an overall transparency indicator (ESG intention) and rank a set of funds in terms of how specific, transparent, and adequate the ESG investment strategy is. A greenwashing sustainable fund may obfuscate its investment strategy by writing mostly vague language whereas a green fund may have a more precise and unambiguous language. To capture the proportion, we propose a scaled ratio metric which is called Language score in this paper. In the pipeline (Figure 1), after obtaining the number of Specific, Ambiguous, and Generic ESG sentences using the best performing model with PEFT, the system calculates the ESG Language score for each document as follows:

$$ESG_{LR} = (X_S / X_A) * X_{SF} \quad (1)$$

where, $X_S$: no. of Specific ESG sentences

$X_A$: no. of Ambiguous ESG sentences

$X_{SF}$: Scaling Factor

It is noteworthy that the ratio ($X_S / X_A$) is a metric which can take following values:

{

  > 1: when $X_S > X_A$ (signifies a greater occurrence of specific ESG language in the fund),

  1: when $X_S == X_A$,

  < 1: when $X_S < X_A$ (signifies a greater occurrence of ambiguous ESG language in the fund), and

  0: when $X_S = 0$

}

$X_{SF}$ is the multiplicative factor, purposefully created to scale the ratio metric ($X_S / X_A$) by constant depending on range of the frequency of specific instances. To tackle the issue of ratio metric becoming undefined, we keep conditional check for $X_A$.

As we establish a scoring method for sustainable funds, we can rank all funds across asset managers and utilize the information for various application e.g., asset allocation, product recommendation. Language rating pertains to creating groups in fund universe to differentiate between clearly written disclosures and vaguely written disclosures. Language rating are derived by quintiling the language scores.

## IV. DATASET PREPARATION

Data is the essential aspect of this research. As mentioned in Step 2 and Step 3 of Figure 1 (Section 3.1), An iterative annotation process was adopted for creating enough samples for training two classifiers namely 1. ESG vs non-ESG model 2. Text clarity classifier. This section describes the data collection process and annotation tasks.

*A. Data Collection*

The process begins with downloading the 'Summary Prospectuses' from literature sections of the official websites of various Asset Management Companies (AMCs). We collected approximately 250 sustainable products prospectuses. For the ESG vs non-ESG model, a rule-based model was employed to identify the ESG and non-ESG sentences which utilizes a ESG diction. We collected 7217 instances of such sentences with 1353 ESG instances and 5863 non-ESG instances. Further, multiple models were experimented to classify ESG vs non-ESG classifier based on sentences collected and annotated from rules-based technique. This lays the process for training ESG vs Non ESG ML classifier (Section 5.2 for performance). The ESG sentences detected by this model is input to the language transparency classier which we discuss next.

With the ESG sentences extracted from 'Principal Investment Strategy' sections of the documents using ESG vs non-ESG classifier, we create a novel 1,155-instance dataset. To the best of our knowledge, this is the first ESG language dataset obtained from fund documents that has been manually annotated for ambiguity. Initially, around 20 sentences from complete set of ESG sentences were then manually annotated for three mutually exclusive labels each: Specific, Ambiguous, and Generic. The labelled instances are used to train the following three sentence transformer models with SetFit: miniLM-v2, paraphrase-mpnet-v2, and sentence-t5-base. The best performing model is then used to obtain weak labels for the remaining (unlabeled) instances in our dataset of ESG sentences. We assess all weak labels and manually correct the incorrectly labeled ones and iterate the process described in Step 3 (Section 3.1). Specific sentences consisted of well-defined rules and inclusion and exclusion criteria regarding the investment strategies, whereas Ambiguous contained sentences that could be argued to have more than one clear understanding or were vague or unclear in terms of their intention. We encountered sentences that were neither Specific nor Ambiguous, such as simple assertive sentences or definitions of terms, which we classified under the Generic category. Some examples of these classes are presented in Table I.

Table I contains examples from our dataset for each of the three mutually exclusive classes: Specific, Ambiguous, and Generic ESG sentences.

*B. Annotation*

Since the use-case is focused only on sustainable investing, the annotators were asked to consider the sentences from the viewpoint of sustainable investment strategy only. The selected and open-sourced dataset was annotated by 3 people with adequate knowledge of ESG investing and were fluent in English with previous exposure of analysing financial documents. They were required to independently provide one of the three mutually exclusive labels (Specific, Ambiguous, or Generic) to the ESG sentences in our dataset*. The dataset was divided into three subsets and each annotator was allocated 2 subset of sentences and was given few weeks to label the sentences. Consequently, each of the 1155 instances was annotated by 2 annotators. Following were the guidelines for annotation.

TABLE I. SAMPLE SENTENCES FROM DATASET

| Specific ESG Sentences |
|---|
| Companies engaged in the business of controversial weapons or that own 25% or more of a company engaged in this activity. |
| The Sub-fund invests a minimum of 5% in green, social, sustainable, and/or sustainability-linked bonds. |
| **Ambiguous ESG Sentences** |
| The Fund will seek to avoid investing in companies that have significant and direct involvement in the manufacturing of alcoholic beverages or gambling. |
| The Fund will seek to invest in companies with sustainable business models which have a strong consideration for ESG risks and opportunities. |
| **Generic ESG Sentences** |
| The ESG Scores used in the S&P SmallCap 600 ESG Index are calculated by the Index Provider. |
| The social pillar factors include workforce, community, product responsibility, and human rights. |

- Choose label deterministically and consistently. Few accurate examples were provided for each label to ensure that understanding of the classes is thorough.
- Specific sentences regarding ESG investment strategy were clear exclusion criteria, clear inclusions, security selection ESG criteria, ESG score rules, specific compliance statements, specific benchmark comparative statements, complying with known standards etc. (Recommended but not limited).
- Ambiguous sentences relating to ESG investment strategy included (but not limited to) ambiguous thresholds, vague sentences, inadequate explanation, proprietary rating systems, fund manager discretions, intentionally obfuscated statements, subjective screening etc.
- Generic statements were mostly factual statement, definitions, or ESG sentences with no investment strategy intention in it. Statement with ESG risk related statements were also separately labelled as Risk category.
- Advised to not make assumption or speculate conclusion based on prior knowledge. Annotators routinely discussed nuances of labelled sentences to make an informed decision and avoid confusion.
- Mark sentences as 'NA' if it is difficult to judge among specific, ambiguous, generic or risk labels.

It can be inferred that the annotators clearly understood the guidelines since in most cases they arrived at the same annotations. Calculating the inter-annotator agreement reveals the degree to which annotators have provided same labels to similar data instances, which reinforces trustworthiness of the data labels. We release gold standard dataset of sentences with agreement on labels by annotators [15].

## V. EXPERIMENTS AND RESULTS

### A. Experiments

We perform experiments and show results on both dataset namely ESG vs non-ESG dataset (7K samples) and text clarity dataset (1155 samples). The transformers and PEFT library from Hugging Face provides multiple latest parameter efficient approaches, and we particularly make of Prompt Tuning approach to report macro F1 scores. Experiments were performed using Amazon EC2 instance, and a single training run of the Prompt Tuning set of experiments took 15 minutes on the dataset. We report macro F1 scores on our dataset. For the train/validation/test splits, we chose a uniform 80% / 10% / 10% to perform the experiments.

### B. Results

In this section, we report different experimental results using large language models. The Table II shows the ESG vs non-ESG classifier results based on machine learning models.

TABLE II. ESG VS NON-ESG Classification (ML MODELS)

| Model | Accuracy | Precision | Recall | F1 |
|---|---|---|---|---|
| Logistic Regression CV | 0.99 | 0.97 | 0.98 | 0.98 |
| Linear SCV | 0.99 | 0.97 | 0.98 | 0.97 |
| Random Forest | 0.99 | 0.93 | 0.99 | 0.96 |
| Multinomial Naïve Bayes | 0.95 | 0.75 | 0.96 | 0.84 |
| Complement Naïve Bayes | 0.95 | 0.92 | 0.84 | 0.84 |

Secondly, we report results with text clarity datasets (1155 instances) using the SetFit finetuning obtained on three sentence transformer models, namely miniLM-v2, paraphrase-mpnet-v2, and sentence-t5-base, in Table III. According to the results, miniLM-v2 gives the overall best performance with F1 score of 0.85, followed by paraphrase-mpnet-v2 and sentence-t5-base.

TABLE III. SENTENCE TRANSFORMER MODEL FINETUNING RESULTS

| Model | Accuracy | Precision | Recall | F1 |
|---|---|---|---|---|
| MiniLM-v2 | 0.85 | 0.87 | 0.85 | 0.85 |
| Paraphrase-mpnet-v2 | 0.85 | 0.86 | 0.85 | 0.85 |
| Sentece-t5-base | 0.82 | 0.85 | 0.83 | 0.83 |

TABLE IV. PROMPT TUNING RESULTS USING BERT MODELS

| Model | Accuracy | Precision | Recall | F1 |
|---|---|---|---|---|
| Deberta-base | 0.83 | 0.85 | 0.83 | 0.84 |
| Roberta-base | 0.83 | 0.83 | 0.83 | 0.83 |
| Bert-base | 0.79 | 0.81 | 0.79 | 0.80 |

Next, we report the Prompt Tuning results on text clarity datasets with three BERT-family models: bert-base, roberta-base, and deberta-base in Table IV. We observe that deberta-base performed the best with 0.84 F1 score, followed by roberta-base and bert-base.

*C. Error Analysis*

In Table V, we see some examples of misclassifications made by our best performing model, i.e., deberta-base fine-tuned with Prompt Tuning PEFT approach. Examples 1 and 4 are 'Specific' but incorrectly labeled by the model, which may be due to the presence of 'up to' and 'better' in the two sentences respectively, which might have led to the model predicting them as 'Ambiguous'. Example 2 is predicted 'Ambiguous' by the model, whereas in fact it is 'Generic'. This can be attributed to the use of the phrase "relevant ESG issues" in the sentence without further expansion on what issues are considered relevant. Example 3 is predicted 'Specific', which could be because the sentence consists of an inclusion criterion.

TABLE V.   MISSCLASSIFICATIONS BY FINETUNED MODEL

| Sentence | Gold Label | Model Prediction |
|---|---|---|
| Up to 3% of the Fund's net assets may be invested in High Social Impact Investments. | Specific | Ambiguous |
| The Fund has access to this research and considers relevant ESG issues. | Generic | Ambiguous |
| Good governance practices include sound management structures, employee relations, remuneration of staff and tax compliance. | Ambiguous | Specific |
| The Sub-fund's weighted average ESG score is better than that of the Paris Aligned Benchmark. | Specific | Ambiguous |

*D. Comparison with Prompt Engineering*

The argument with recent generative models like GPT 4 models is that such classification may be possible as it is instruction finetuned with thousands of tasks and millions of data points from human feedbacks. We experimented with prompts to GPT 3.5 turbo and GPT-4 to achieve the same task of clarity classification as reported in Table VI.

TABLE VI.   PROMPT BASED CLASSIFICATION RESULTS

| Model | Accuracy | Precision | Recall | F1 |
|---|---|---|---|---|
| GPT-3.5 Turbo | 0.57 | 0.58 | 0.56 | 0.53 |
| GPT-4 | 0.68 | 0.74 | 0.70 | 0.70 |

We could engineer prompt better than the above results and believe that such in-domain and complex classification task can not be accurately solved by pretrained generative LLMs like GPT-4. Fine-tuning remains the best approach to follow.

VI. CONCLUSION

This paper attempts to address the ongoing issue of understanding how compliant and accurately a sustainable fund reflects its ESG strategy by measuring sustainability intention. The paper proposes a language metric as a transparency indicator and utilizes it to rank the language of ESG disclosures. We established a framework employing multiple models to surface specific and vague ESG related languages. This also demonstrate utilizing various finetuning techniques in a scarce labelled data scenario. Few shot finetuning a smaller model (in order of hundreds) outperforms zero-shot prompting techniques by more than 10%-15% in F1 score. Overall, the solution ranks sustainable universe with language score and may assist advisors and investors in screening of funds. We present this work to the NLP research community as a jumpstart towards more rigorous experiments in the ESG domain. We published a dataset "ESG-Prospectus-Clarity-Category" to support further research. Any data produced during this study are released publicly for further research by the community under CC-BY-NC-SA 4.0.

As future work, we intend to extend our work to understand prospectus language in terms of exclusion categories (anti-ESG business), ESG themes and compliance checks to investigate and compare insights from sustainable fund universe.

## A  APPENDICES

In the appendix section, we provide additional details which could not be included in the main paper.

TABLE VII.  Specific(green), Ambiguous(red), and Generic(black) ESG sentences detected by our model in the Summary Prospectuses of the following fund.

**BLACKROCK FUTURE CLIMATE AND SUSTAINABLE ECONOMY ETF**

The Fund seeks to maximize total return by primarily investing in companies that BFA believes are furthering the transition to a lower carbon economy including themes such as sustainable energy, circular economy, future of transport and nutrition. The Fund will also seek to invest in a portfolio of equity securities that, in BFA's view, has an aggregate environmental assessment that is better than the aggregate environmental assessment of the MSCI ACWI Multiple Industries Select Index (the "MSCI ACWI Select" or the "Underlying Index").

The Fund may invest in equity securities issued by U.S. and non-U.S. companies in any market capitalization range.

**iSHARES ESG AWARE MSCI USA ETF**

The Index Provider begins with the Parent Index and excludes securities of companies involved in the business of tobacco, companies involved with controversial weapons, producers and retailers of civilian firearms, and companies included in certain fossil fuels-related activity such as the production of thermal coal, thermal coal-based power generation and extraction of oil sands based on revenue or percentage of revenue thresholds for certain categories (e.g., $20 million or 5%) and categorical exclusions for others (e.g., controversial weapons).

The Index Provider also excludes companies that are directly involved in very severe, ongoing business controversies (in each case as determined by the Index Provider), and then follows a quantitative process that is designed to determine optimal weights for securities to maximize exposure to securities of companies with higher ESG ratings, subject to maintaining risk and return characteristics similar to the Parent Index.

For each industry, the Index Provider identifies key ESG issues that can lead to unexpected costs for companies in the medium- to long-term.